\documentclass[10pt,twocolumn,letterpaper]{article}

\usepackage{wacv}
\usepackage{times}
\usepackage{epsfig}
\usepackage{graphicx}
\usepackage{amsmath}
\usepackage{amssymb}
\usepackage{booktabs}

\def\wacvPaperID{70} 

\wacvalgorithmstrack   

\wacvfinalcopy 

\ifwacvfinal
\usepackage[breaklinks=true,bookmarks=false,colorlinks]{hyperref}
\else
\usepackage[pagebackref=true,breaklinks=true,colorlinks,bookmarks=false]{hyperref}
\fi

\usepackage{booktabs}
\usepackage{tabularx}
\usepackage{mathtools}

\usepackage{color, colortbl}
\usepackage[normalem]{ulem}
\usepackage{subcaption}
\newcommand{\parsection}[1]{\vspace{2mm}\noindent\textbf{#1 }}
\usepackage{multirow}
\makeatletter
\newcommand*\bigcdot{\mathpalette\bigcdot@{0.9}}
\newcommand*\bigcdot@[2]{\mathbin{\vcenter{\hbox{\scalebox{#2}{$\m@th#1\bullet$}}}}}
\makeatother

\usepackage{algorithm}
\usepackage{listings}
\usepackage{etoolbox}
\makeatletter
\AfterEndEnvironment{algorithm}{\let\@algcomment\relax}
\AtEndEnvironment{algorithm}{\kern2pt\hrule\relax\vskip3pt\@algcomment}
\let\@algcomment\relax
\newcommand\algcomment[1]{\def\@algcomment{\footnotesize#1}}
\renewcommand\fs@ruled{\def\@fs@cfont{\bfseries}\let\@fs@capt\floatc@ruled
  \def\@fs@pre{\hrule height.8pt depth0pt \kern2pt}%
  \def\@fs@post{}%
  \def\@fs@mid{\kern2pt\hrule\kern2pt}%
  \let\@fs@iftopcapt\iftrue}
\makeatother
\usepackage[algo2e]{algorithm2e}

\begin{document}

\title{Efficient Visual Tracking with Exemplar Transformers}

\author{Philippe Blatter$^{*,1}$ \quad
	Menelaos Kanakis$^{*,1}$\quad
    Martin Danelljan$^{1}$\quad
	Luc Van Gool$^{1,2}$ \vspace{2mm} \\
$^1$ETH Z\"urich \quad $^2$KU Leuven
}


\maketitle
\thispagestyle{empty}

\begin{abstract}
    The design of more complex and powerful neural network models has significantly advanced the state-of-the-art in visual object tracking.
    These advances can be attributed to deeper networks, or the introduction of new building blocks, such as transformers.
    However, in the pursuit of increased tracking performance, runtime is often hindered. 
    Furthermore, efficient tracking architectures have received surprisingly little attention.
    In this paper, we introduce the Exemplar Transformer, a transformer module utilizing a single instance level attention layer for realtime visual object tracking.
    E.T.Track, our visual tracker that incorporates Exemplar Transformer modules, runs at 47~\textit{FPS} on a CPU.
    This is up to 8$\times$ faster than other transformer-based models.
    When compared to lightweight trackers that can operate in realtime on standard CPUs, E.T.Track consistently outperforms all other methods on the LaSOT~\cite{fan2019lasot}, OTB-100~\cite{wu2013online}, NFS~\cite{kiani2017need}, TrackingNet~\cite{muller2018trackingnet}, and VOT-ST2020~\cite{kristan2020eighth} datasets. Code and models are available at \url{https://github.com/pblatter/ettrack}.
\end{abstract}

{\let\thefootnote\relax\footnote{{$^*$P. Blatter and M. Kanakis contributed equally to this work.}}}

\label{sec:abs}

\section{Introduction}

Estimating the trajectory of an object in a video sequence, referred to as visual tracking, is one of the fundamental problems in computer vision. 
Deep neural networks have significantly advanced the performance of visual tracking methods with deeper networks~\cite{bertinetto2016fully}, more accurate bounding boxes~\cite{li2018high}, or with the introduction of new modules, such as transformers~\cite{stark,transformer_tracker,transformer_meets_tracker}.
However, these advances often come at the cost of more expensive models.
While the demand for realtime visual tracking on applications such as autonomous driving, robotics, and human-computer-interfaces is increasing, efficient deep tracking architectures have received surprisingly little attention.
This calls for visual trackers that, while accurate and robust, are capable of operating in realtime under the hard computational constraints of limited hardware.

\begin{figure}[t]
    \centering%
    \vspace{0.1 in}
	\resizebox{\linewidth}{!}{%
    \includegraphics[width=0.85\textwidth]{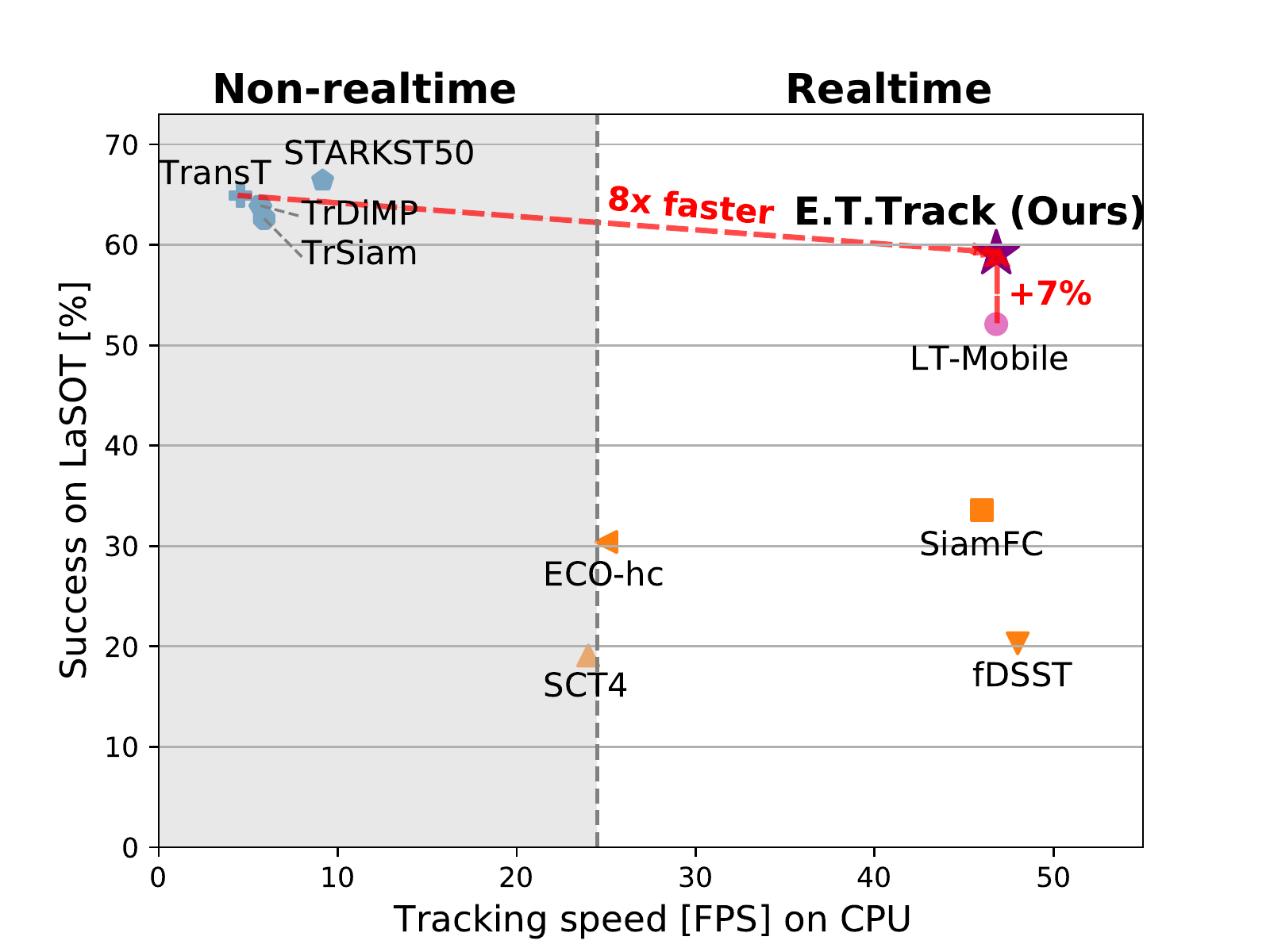}
    
    }
  \caption{Comparison of tracker performance in terms of AUC score (Success in \%) on LaSOT vs.\ tracking speed in \textit{FPS} on a standard CPU. Our Exemplar Transformer Tracker (E.T.Track) outperforms all other realtime trackers. It achieves a $7\%$ higher AUC score than LT-Mobile~\cite{yan2021lighttrack}. Furthermore, our approach achieves up to $8\times$ faster runtime on a CPU compared to previous Transformer-based trackers.
    }
    \label{fig:teaser_figure}
    \vspace{-0.1 in}
\end{figure}

Transformers~\cite{vaswani2017attention}, proposed for machine translation, have also demonstrated superior performance in a number of vision based tasks, including image~\cite{bello2019attention} and video~\cite{wang2018non} classification, object detection~\cite{carion2020end}, and even multi-task learning~\cite{bruggemann2021exploring}.
The field of visual tracking has also observed similar performance benefits~\cite{stark,sun2020transtrack,transformer_meets_tracker}.
While transformers have enabled the trackers to improve accuracy and robustness, they severely suffer from high computational cost, leading to decreased runtime operation, as depicted in Fig.~\ref{fig:teaser_figure}.
In this work, we set out to find a transformer module, capable increasing tracking accuracy and robustness while not compromising runtime. 

In this work we propose Exemplar Attention, a single instance-level attention layer for visual tracking. 
Our attention module exploits domain specific knowledge to improve the tracker's performance, while maintaining a comparable runtime.
Specifically, we build upon two hypotheses. 
Firstly, one global query value is sufficiently descriptive when tracking a single object.
Secondly, a small set of exemplar values can act as a shared memory between the samples of the dataset.
Thus, our constrained instance-level Exemplar Attention captures more explicit information about the target object, compared to conventional attention modules.

We develop the Exemplar Transformer Tracker (E.T.Track) by integrating our Exemplar Transformer layer into a Siamese tracking architecture.
Specifically, we replace the convolutional layers in the tracker heads with the Exemplar Transformer layer. 
The additional expressivity from the Exemplar Transformer layer significantly improves the performance of the models based on regular convolutional layers. 
The added performance gain comes at an insignificant cost in runtime, as seen in Fig.~\ref{fig:teaser_figure} when comparing it to the mobile LightTrack~\cite{yan2021lighttrack} (LT-Mobile).
We further compare our transformer layer for single object tracking to other generic transformer layers.
We find that Exemplar Transformer consistently outperforms competing methods, attesting to the benefits of explicitly designing attention layers for the task of visual tracking.

We validate our approach on six benchmark datasets: LaSOT~\cite{fan2019lasot}, OTB-100~\cite{wu2013online}, UAV-123~\cite{mueller2016benchmark}, NFS~\cite{kiani2017need}, TrackingNet~\cite{muller2018trackingnet} and VOT-ST2020~\cite{kristan2020eighth}.
Our proposed tracker runs at $46.8$~ Frames Per Second (\textit{FPS}) on a CPU, while setting a new state-of-the-art among realtime CPU trackers by achieving $59.1\%$ AUC on the challenging LaSOT dataset.

In summary, our contributions are:
 \begin{itemize}
     \item We introduce Exemplar Transformer, a transformer layer based on a single instance-level attention layer referred to as Exemplar Attention. 
     
     \item We develop a transformer-based tracking architecture based upon our Exemplar Transformer layer. 
     
     \item Our tracker runs in realtime on a CPU, while outperforming previous realtime trackers on 5 benchmarks.
\end{itemize}

\label{sec:intro}

\section{Related Work}

\parsection{Siamese Trackers}
In recent years, Siamese trackers have gained significant popularity due to their performance capabilities and simplicity. 
The Siamese-based tracking framework formulates visual object tracking as a template matching problem, utilizing cross-correlation between a search and an image patch. 
The original work of Bertinetto~\etal introduced SiamFC~\cite{bertinetto2016fully}, the first model incorporating feature correlation into a Siamese framework. 
Li~\etal~\cite{li2018high} introduced region proposal networks to increase efficiency and obtain more accurate bounding boxes. 
More recent advances on the Siamese tracker front include the use of additional branches \cite{wang2019fast}, refinement modules for more precise bounding box regression \cite{yan2021alpha}, and various model update mechanisms \cite{gao2019graph,guo2017learning,yang2018learning,zhang2019learning}. 
Unlike previous Siamese trackers, we propose the Examplar Transformer module that is incorporated into the prediction heads, and improves the tracker's performance at an insignificant runtime increase.

\parsection{Transformers in Tracking}
The Transformer~\cite{vaswani2017attention} was introduced as a module to improve the learning of long-range dependencies in neural machine translation, by enabling every element to attend to all others.
In computer vision, transformers have been used in image~\cite{bello2019attention} and video~\cite{wang2018non} classification, object detection~\cite{carion2020end}, and even multi-task learning of dense prediction tasks~\cite{bruggemann2021exploring}.
More related to our work, transformers have also been utilized to advance the performance of visual trackers. 
STARK~\cite{stark} utilizes transformers to model the global spatio-temporal feature dependencies between target object and search regions. 
This is achieved by integrating a dynamically updated template into the encoder, in addition to the regular search and template patch.
\cite{transformer_meets_tracker} introduced a transformer architecture that improves the standard Siamese-like pipeline by additionally exploiting temporal context. 
The encoder model mutually reinforces multiple template features by leveraging self-attention blocks. 
In the decoder, the template and search branch are bridged by cross-attention blocks in order to propagate temporal contexts.
\cite{transformer_tracker} also improve Siamese-based trackers by replacing the regular correlation operation by a Transformer-based feature fusion network. 
The Transformer-based fusion model aggregates global information, providing a superior alternative to the standard linear correlation operation. 
ToMP~\cite{mayer2022transforming}, on the other hand, utilizes a transformer to predicts the weights of a convolutional kernel in order to localize the target in the search region and template patches.
In this work, we also design transformer architecture for tracking.
Unlike the previous transformers for tracking, Exemplar Transformer is lightweight and can be utilized in computationally limited hardware running at realtime.

\parsection{Efficient Tracking Architectures} 
With an increase in demand for realtime visual tracking in applications such as autonomous driving, and human-computer-interfaces, efficient deep tracking architectures are essential.
Surprisingly, however, little attention has been provided on efficient trackers that can operate on computationally limited hardware.
KCF~\cite{henriques2014high} and fDSST~\cite{danelljan2016discriminative} employ hand-crafted features to enable realtime operation on CPUs.
While fast, their reliance on hand crafted features significantly hinders their performance compared to newer and more complex methods.
In contrast, we present an efficient deep tracker that operates at a comparable runtime but performs on par with the more expensive deep trackers.
More related to out work, LightTrack~\cite{yan2021lighttrack} employs neural architecture search (NAS) to find a lightweight and efficient Siamese tracking architecture. 
We instead propose an efficient transformer layer that can complement existing architecture advances such as LightTrack.
Specifically, our transformer layer can act as a drop in replacement for convolutional layers, increasing performance with negligible effects on runtime.

\begin{figure}[t]
\centering%
	\resizebox{\linewidth}{!}{%
    \includegraphics[width=0.85\textwidth]{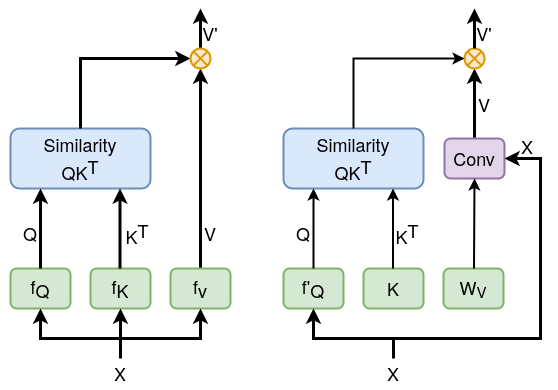}
    }
  \caption{Comparison of our Exemplar Attention module (right) and the standard scaled dot-product attention module~\cite{vaswani2017attention}. The matching blocks are indicated by identical colours. The line thickness is indicative of the tensor size.}
    \label{fig:att_module}
\end{figure}

\parsection{Efficient Transformers}
The immense interest in transformer architectures \cite{dosovitskiy2020image,parmar2018image,carion2020end,wang2018non} resulted in the development of various efficient model variants that can be grouped in 4 major categories~\cite{tay2020efficient}.
\emph{Low Rank/ Kernel} methods assume and leverage low-rank approximations of the self-attention matrix~\cite{katharopoulos2020transformers,choromanski2020masked}.
\emph{Memory/ Downsampling} methods learn a side memory module to access multiple tokens simultaneously, or simply reduce the sequence length~\cite{tang2022patch,zhang2021rest,liu2021swin}.
\emph{Fixed/ Factorized/ Random Patterns} limit the field of view of the self-attention, such as using block patterns~\cite{ryoo2021tokenlearner,zhang2021rest,liu2021swin,ramachandran2019stand}.
\emph{Learnable Patterns} replace the fixed pattern, as in standard transformers, with dynamic patterns~\cite{vyas2020fast,wang2020cluster,kitaev2020reformer}.
Our work falls in the intersection of \emph{Memory/ Downsampling} and \emph{Fixed/ Factorized/ Random Patterns}.
Unlike the aforementioned works that aim to design generic attention layers, Exemplar Attention is instead designed for the task of single object visual tracking by exploiting domain specific knowledge.

\label{sec:related_work}

\section{Efficient Tracking with Transformers}
\label{sec:eff_tracker}

Striking a balance between well performing object trackers and runtime speeds that fall in the realtime envelope, is a challenging problem when deploying on computationally limited devices.
In this section, we introduce the Exemplar Transformer, a transformer architecture based on single instance level attention layers for single object tracking.
While lightweight, our Exemplar Transformer significantly closes the performance gap with the computationally expensive transformer-based trackers~\cite{stark,transformer_meets_tracker,transformer_tracker}.
Sec.~\ref{sec:eff_trans} first presents the original Transformer of Vaswani~\etal~\cite{vaswani2017attention}, followed by our Exemplar Transformer formulation.
Sec.~\ref{sec:archit_over} introduces our E.T.Track. 
Specifically, it first outlines the overall architecture, and presents how Exemplar Transformers are utilized within the tracker.

\subsection{Exemplar Transformers}
\label{sec:eff_trans}

\parsection{Standard Transformer}
The Transformer~\cite{vaswani2017attention}, introduced for machine translation, receives a one dimensional input sequence $x \in \mathbb{R}^{N \times D}$ with $N$ feature vectors of dimensions $D$.
The input sequence is processed by a series of transformer layers defined as
\begin{equation}
    \label{eq:transformer}
    T(x) = f(A(x) + x).
\end{equation}
The function $f(\cdot)$ is a lightweight Feed-Forward Network (FFN) that projects independently each feature vector.
The function $A(\cdot)$ represents a self-attention layer that acts across the entire sequence.
Specifically, the authors used the ``Scaled Dot-Product Attention'', defined as
\begin{equation} 
\label{eq:sdpa}
\begin{split}
    A(x) & = \text{softmax}\Bigg(\frac{\overbrace{Q \vphantom{K^T}}^{f_Q(x)} \overbrace{K^T}^{f_K(x)}}{\underbrace{\sqrt{d_k}}_{\text{constant}}}\Bigg) \overbrace{V}^{f_V(x)} \\
        & = \text{softmax}\Bigg(\frac{\overbrace{(xW_Q) \vphantom{W_K^Tx^T}}^{f_Q(x)} \overbrace{(W_K^Tx^T)}^{f_K(x)}}{\underbrace{\sqrt{d_k}}_{\text{constant}}}\Bigg) \overbrace{(xW_V)}^{f_V(x)}.
\end{split}
\end{equation}
The queries $Q \in \mathbb{R}^{N \times D_{QK}}$, keys $K \in \mathbb{R}^{N \times D_{QK}}$, and values $V \in \mathbb{R}^{N \times D_V}$ represent projections of the input sequence, while $\sqrt{d_k}$ is a normalization constant. 
The self attention, therefore, computes a similarity score between all representations, linearly combines the feature representations, and accordingly adapts the input representation $x$ in Eq.~\ref{eq:transformer}. 
The computational complexity of Eq.~\ref{eq:sdpa} is $\mathcal{O}(N^2 D)$, \ie it scales quadratically with the length of the input sequence.

\parsection{Exemplar Attention}
We now introduce Exemplar Attention, the key building block of the Exemplar Transformer module.
We hypothesize that, while the direct connection between all features is essential in machine translation and some vision tasks, this design choice can be sub-optimal when attending to the single object being tracked.
We describe the required modifications of the individual components below.

The standard Query function $f_Q$ projects every spatial location of the feature map independently to a query space.
Unlike machine translation where every feature represents a specific word or token, adjacent spatial representation in vision tasks often correspond to the same object.
Consequently, we aggregate the information of the feature map $X \in \mathbb{R}^{H \times W \times D}$, where $H \times W$ represents the spatial dimensions.
Specifically, we use a 2D adaptive average pooling layer with an output spatial dimension $S\times S$, followed by a flattening operation.
The operation is denoted $\Psi_S(X)$, decreasing the output spatial dimension to $S^2$. 
The compressed representation of $X$ is then projected to a query space as in the standard self-attention formulation. 
\begin{equation} \label{eq:avg_pool}
     Q = \Psi_S(X)W_Q \in \mathbb{R}^{S^2 \times D_{QK}}
\end{equation}
We hypothesize that for single instance tracking, one global query value
is sufficient to identify the object of interest, while also decreasing the computational complexity of the module.
To this extent, we set $S=1$.
This design choice is further supported by the success of global pooling in classification architectures~\cite{he2016deep}, as well as transformer based object detection~\cite{carion2020end}.

The keys and values, as presented in Eq.~\ref{eq:sdpa}, are per spatial location linear projections of the input.
The self-attention layer then enables the learning of spatial correlations, at the cost of every feature attending to all others.
This eliminates spatial biases built into convolutional layers.
Rather than requiring a fine grained feature map and relying solely on intra-sample relationships, we instead learn a small set of exemplar representations.
The exemplar representations encapsulates dataset information in order to dynamically adapt the attention layer given the global query token and the captured information.
To this end, we optimize a small set of exemplar keys $K = \hat{W}_K \in \mathbb{R}^{E \times D_{QK}}$ that, unlike the formulation in Eq.~\ref{eq:sdpa}, are independent of the input.
The similarity matrix therefore associates the global query, Eq.~\ref{eq:avg_pool}, to exemplars.
Our attention layer then refines the input representation on the local level by replacing the projection $f_V(\cdot)$ with a convolutional operation
\begin{equation} \label{eq:exemplar_conv}
     V = W_V \circledast X \in \mathbb{R}^{E \times H \times W  \times D_V},
\end{equation}
where $W_V \in \mathbb{R}^{E \times Z \times Z}$ can be of any spatial dimension $Z$, while the number of exemplars $E$ can be chosen arbitrarily.
We use $E=4$ in our experiments, which is significantly smaller than the dimensions $H \times W$, maintaining comparable runtime.

\begin{figure}[t]
    \centering
    \vspace{0.2in}
    \includegraphics[width=\linewidth]{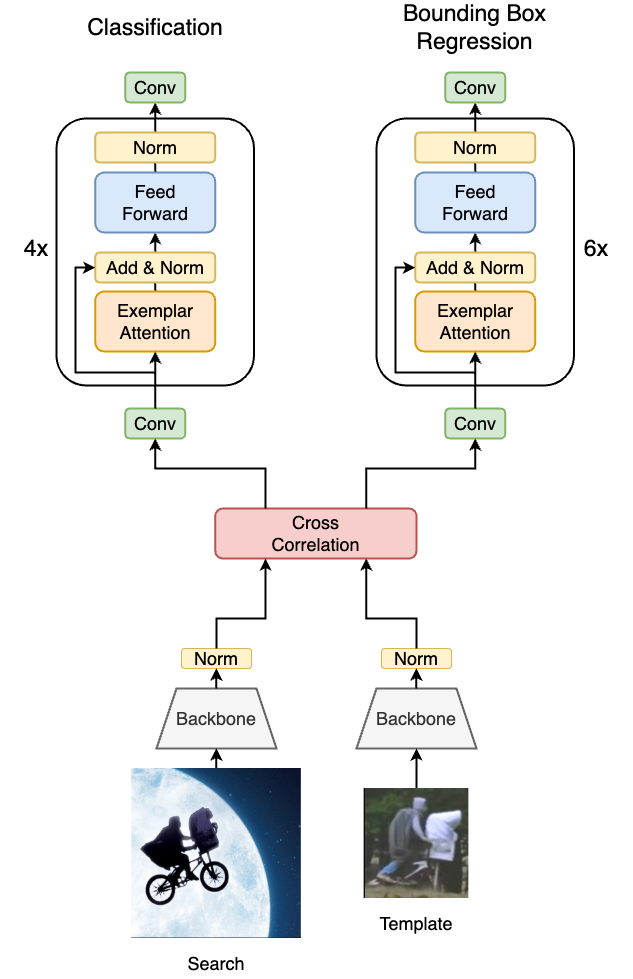}
    \caption{E.T.Track - a Siamese tracking pipeline that incorporates Exemplar Transformers in the tracker head.}
    \label{fig:tracker}
\end{figure}

Our efficient Exemplar Attention is therefore defined as,
\begin{equation} 
\label{eq:et-final}
  A(x) = \text{softmax}\Bigg(\frac{\overbrace{(\Psi_S(X)W_Q) \vphantom{\hat{W}_K^T}}^{f_Q(x)} \overbrace{(\hat{W}_K^T)}^{f_K(\cdot)}}{\underbrace{\sqrt{d_k}}_{\text{constant}}}\Bigg) \overbrace{(W_V \circledast X)}^{f_V(x)},
\end{equation}
but can also be written as,
\begin{equation} 
\label{eq:et-final_light}
  A(x) = \Bigg[\text{softmax}\Bigg(\frac{(\Psi_S(X)W_Q) (\hat{W}_K^T)}{\sqrt{d_k}}\Bigg) W_V\Bigg] \circledast X. 
\end{equation}

Exemplar Attention, while inspired by the scaled dot-product attention, is conceptually very different. 
In self-attention \eqref{eq:sdpa}, $f_{\{Q,K,V\}}$ act as projections to their corresponding feature spaces, with the similarity function learning the relationships between all spatial locations.
In other words, self-attention relies solely on intra-sample relationships, and therefore requires fine-grained representations.
Instead, the Exemplar Attention layer enforces the attending over a single instance through the use of a global query token.
The global query encapsulated the representation of the object, is dynamically generated from the input image, and applied locally on the feature map using a convolutional operation.
To enable the use of a single query token, we exploit dataset information to form the exemplar representations through end-to-end optimization, eliminating the need of the intra-sample similarity function.
A comparison between the two attention mechanisms is depicted in Fig.~\ref{fig:att_module}.

\subsection{E.T.Track Architecture}
\label{sec:archit_over}

In this section we introduce the base tracking architecture used throughout our work.
While Exemplar Transformers can be incorporated into any tracking architecture, we evaluate its efficacy on lightweight Siamese trackers.
An overview of the E.T.Track architecture can be seen in Fig.~\ref{fig:tracker}.

Our model employs the lightweight backbone model LT-Mobile~\cite{yan2021lighttrack}. 
The model was identified by NAS on a search space consisting of efficient and lightweight building blocks. 
The feature extracting backbone consists of $3\times 3$ convolutional layers, depthwise separable convolutional layers and mobile inverted bottleneck layers with squeeze and excitation modules. 

The Exemplar Transformer layer can act as a drop in replacement for any convolution operation of the architecture. 
We replace all the convolutions in the classification and bounding box regression branches, while keeping the lightweight backbone architecture untouched. 
This eliminates the need for retraining the backbone on ImageNet~\cite{deng2009imagenet}.

Search and template frames are initially processed through a backbone network. 
The similarity between the representations is computed by a pointwise cross-correlation. 
The resulting correlation map is then fed into the tracker head, where it is processed by a classification branch and a bounding box regression branch in parallel. 
The bounding box regression branch predicts the distance to all four sides of the bounding box. 
The classification branch predicts whether each region is part of the foreground or the background. 
During training, the bounding box regression branch considers all the pixels within the ground truth bounding box as training samples, therefore, the model is able to determine the exact location of the object even when only small parts of the input image are classified as foreground. 
The model is trained by optimizing a weighted combination of the binary cross-entropy (BCE) loss and the IoU loss~\cite{yu2016unitbox} between the predicted and ground-truth bounding boxes. 
For more details, as well as more information on the data preprocessing, we refer the reader to \cite{zhang2020ocean}.

\label{sec:method}

\section{Experiments}

\definecolor{Gray}{gray}{0.9}
\newcolumntype{g}{>{\columncolor{Gray}}c}

\begin{figure}
    \vspace{0.1in}
    \begin{center}
        \includegraphics[width=0.9\linewidth]{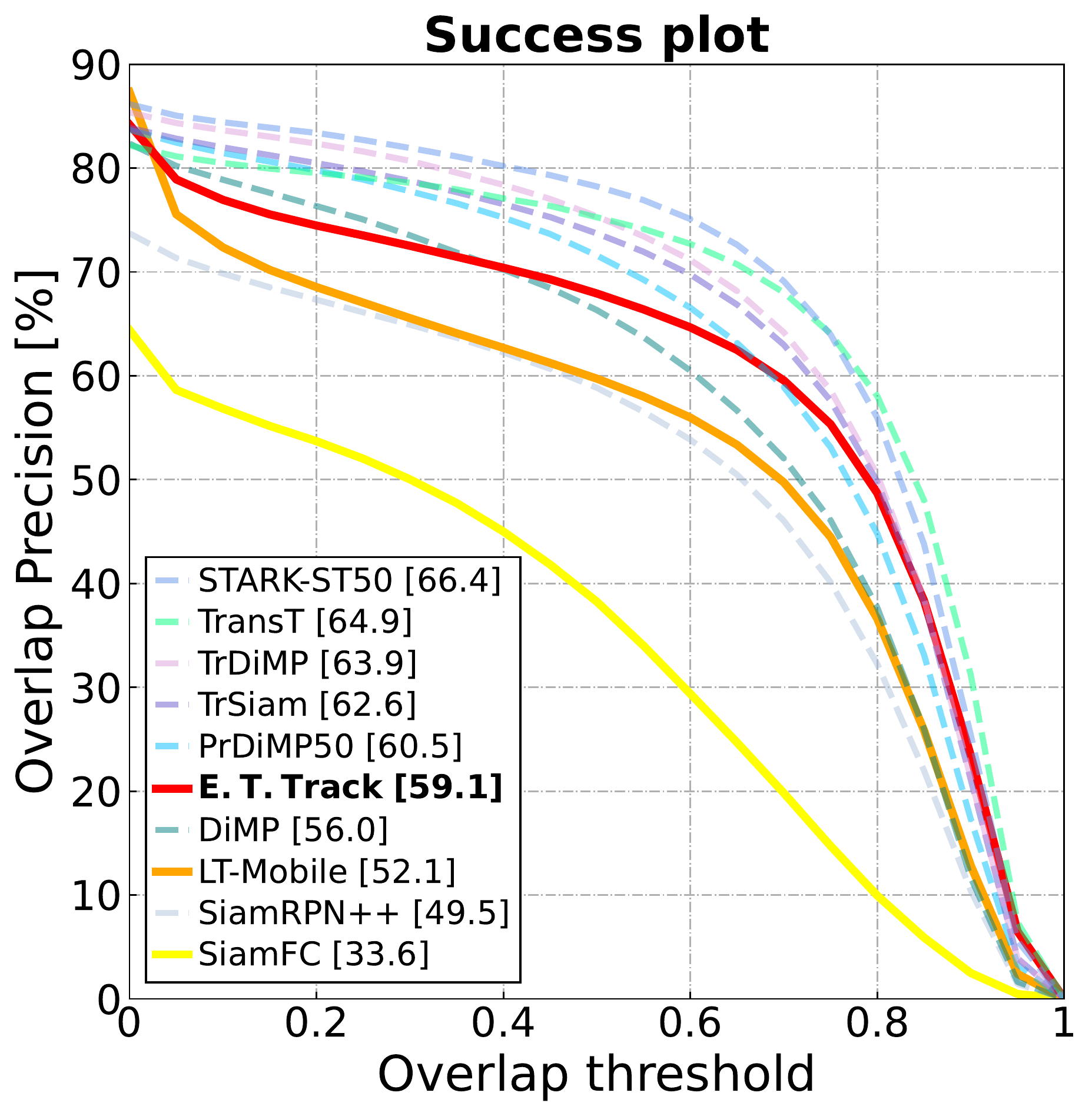}
        \caption{Success plot on the LaSOT dataset. The CPU realtime trackers are indicated by continuous lines in warmer colours, while the non-realtime trackers are indicated by dashed lines in colder colours. 
        E.T.Track significantly outperforms the other realtime trackers, and even outperforms some of the more established trackers such as DiMP~\cite{bhat2019learning}.
        Furthermore, it significantly closes the performance gap with the more expensive transformer trackers.}
        \label{fig:lasot_success}
    \end{center}
    \vspace{-0.2 in}
\end{figure}

\begin{table*}[t!]
  \centering
\scalebox{0.7}{
  \begin{tabular}{@{}cgggggggggccc@{}}
    \toprule
    & \multicolumn{9}{g}{non-realtime} & \multicolumn{3}{c}{realtime} \\
    \midrule
    & ATOM & SiamRPN++ & DiMP-50 & PrDiMP-50 & SiamR-CNN & TransT & TrDiMP & TrSiam & STARK-ST50 & ECO & LT-Mobile & \textbf{E.T.Track}  \\
    & \cite{danelljan2019atom} & \cite{li2019siamrpn++} & \cite{bhat2019learning} & \cite{danelljan2020probabilistic} & \cite{voigtlaender2020siam} & \cite{transformer_tracker} & \cite{transformer_meets_tracker} & \cite{transformer_meets_tracker} & \cite{stark} & \cite{danelljan2017eco} & \cite{yan2021lighttrack} & \textbf{(Ours)} \\
    \midrule
    NFS & 58.4 & 50.2 & 62 & 63.5 & 63.9 & 65.7 & \textcolor{blue}{66.5} & 65.8 & 66.4 & 46.6 & 55.3 &  \textbf{\textcolor{red}{59.0}} \\
    UAV-123 & 64.2 & 61.3 & 65.3 & 68 & 64.9 & \textcolor{blue}{69.4} & 67.5 & 67.4 & 68.8 &  51.3 & \textbf{\textcolor{red}{62.5}} & 62.3 \\
    OTB-100 & 66.9 & 69.6 & 68.4 & 69.6 & 70.1 & 69.1 & \textcolor{blue}{71.1} & 70.8 & 67.3 & 64.3 & 66.2 & \textbf{\textcolor{red}{67.8}} \\
    \midrule
    CPU Speed  & 20 & 15 & 15 & 15 & 15 & 5 & 6 & 6 & 9  & 25 & 47 & 47 \\
    \bottomrule
  \end{tabular}}
  \caption{State-of-the-art comparison on the NFS, OTB-100 and UAV-123 datasets in terms of Area Under the Curve (AUC). The best score is highlighted in \textcolor{blue}{blue} while the best realtime score is highlighted in \textcolor{red}{red}. We additionally report CPU runtime speeds in \textit{FPS}.} 
 \label{tab:uav_nfs_otb}
\end{table*}

\begin{table*}[t]
  \centering
\scalebox{0.7}{
  \begin{tabular}{@{}cgggggggggccc@{}}
    \toprule
    & \multicolumn{9}{g}{non-realtime} & \multicolumn{3}{c}{realtime} \\
    \midrule
    & ATOM & SiamRPN++ & DiMP-50 & PrDiMP-50 & SiamR-CNN & TransT &  TrDiMP & TrSiam &STARK-ST50 & ECO & LT-Mobile & \textbf{E.T.Track}  \\
    & \cite{danelljan2019atom} & \cite{li2019siamrpn++} & \cite{bhat2019learning} &  \cite{danelljan2020probabilistic} & \cite{voigtlaender2020siam} & \cite{transformer_tracker} &  \cite{transformer_meets_tracker} & \cite{transformer_meets_tracker} & \cite{stark} & \cite{danelljan2017eco} &  \cite{yan2021lighttrack} & \textbf{(Ours)}\\
    \midrule
    Prec. (\%) & 64.84 & 69.38 & 68.7 & 70.4 & 80 & \textcolor{blue}{80.3} & 73.1 & 72.7 &  - & 48.86  &  69.5 & \textbf{\textcolor{red}{70.6}} \\
    N. Prec. (\%)  & 77.11 & 79.98 & 80.1 & 81.6 & 85.4 & \textcolor{blue}{86.7} & 83.3 & 82.9 & 86.1 & 62.14 & 77.9 & \textbf{\textcolor{red}{80.3}}  \\
    Success (\%) & 70.34 & 73.3 & 74 & 75.8 & 81.2 & \textcolor{blue}{81.4} & 78.4 & 78.1 & 81.3 & 56.13 & 72.5 & \textbf{\textcolor{red}{75.0}} \\
    \midrule
    CPU Speed  & 20 &  15 & 15 & 15 & 15 & 5 & 6 & 6 & 9  & 25 & 47 & 47 \\
    \bottomrule
  \end{tabular}}
  \caption{State-of-the-art comparison on the TrackingNet test set, comprised of 511 sequences. 
  The trackers are compared in terms of Precision (Prec.), Normalized Precision (N. Prec.), and Success.
  The best score is highlighted in \textcolor{blue}{blue} while the best realtime score is highlighted in \textcolor{red}{red}. We additionally report CPU runtime speeds in \textit{FPS}.} 
  
 \label{tab:trackingnet}
\end{table*}

\begin{table*}[t]
  \centering%
\scalebox{0.8}{
  \begin{tabular}{@{}cgggggccc@{}}
    \toprule
    & \multicolumn{5}{g}{non-realtime} & \multicolumn{3}{c}{realtime} \\
    \midrule
    & SiamFC & ATOM & DiMP & SuperDiMP & STARK-ST50 & KCF & LT-Mobile & E.T.Track \\
    &  \cite{bertinetto2016fully} & \cite{danelljan2019atom} & \cite{bhat2019learning} & \cite{superdimp} & \cite{stark} & \cite{henriques2014high} &  \cite{yan2021lighttrack} & \textbf{(Ours)} \\
    \midrule
    EAO & 0.179 & 0.271 & 0.274 & 0.305 & \textcolor{blue}{0.308} & 0.154 &  0.242 & \textbf{\textcolor{red}{0.267}}\\
    Accuracy & 0.418 & 0.462 & 0.457 & 0.477 & \textcolor{blue}{0.478} & 0.407 & 0.422 & \textbf{\textcolor{red}{0.432}}\\
    Robustness & 0.502 & 0.734 & 0.740 & 0.786 & \textcolor{blue}{0.799} & 0.432 & 0.689 & \textbf{\textcolor{red}{0.741}}\\
    \midrule
    CPU Speed  & 6 & 20 & 15 & 15 & 9  & 95 & 47 & 47 \\\bottomrule
  \end{tabular}}
  \caption{Comparison of bounding box predicting trackers on the VOT-ST2020 dataset. 
  We report the Expected Average Overlap (EAO), the Accuracy and Robustness.
  The best score is highlighted in \textcolor{blue}{blue} while the best realtime score is highlighted in \textcolor{red}{red}. We additionally report CPU runtime speeds in \textit{FPS}.} 
 \label{tab:vot2020}
\end{table*}

We first present implementation details of our tracker in section \ref{sec:impl_details}. 
The comparison to state-of-the-art is presented in Sec.~\ref{sec:results}, followed by an ablation study in Sec.~\ref{sec:ablation}.
Code and trained models will be released on publication.

\subsection{Implementation Details}
\label{sec:impl_details}
\parsection{Architecture} 
We adopt the LT-Mobile architecture of LightTrack~\cite{yan2021lighttrack} as our baseline due to its performance versus efficiency trade-off.
LT-Mobile is comprised of a small encoder, and later branches to the classification and regression heads.
The classification head is comprised of 6 convolutional modules, while the regression head is comprised of 8.
Each convolutional module consists of a Depthwise Separable Convolution~\cite{howard2017mobilenets}, a Batch Normalization layer~\cite{ioffe2015batch} and a Rectified Linear unit.
E.T.Track replaces each convolutional module with an Exemplar Transformer Layer, introduced in Sec.~\ref{sec:eff_trans}.
The learnable ``value'' parameters of the attention module are initialized using kaiming initialization~\cite{he2015delving}, while the learnable ``keys'' parameters are initialized using a normal distribution. 
The FFN consists of 2 linear layers with a ReLU activation, dropout~\cite{srivastava2014dropout} with a ratio of 0.1, and LayerNorm~\cite{ba2016layer}.

\parsection{Training}
All models have been trained using an Nvidia GTX TITAN X, and evaluated on an Intel(R) Core(TM) i7-8700 CPU @ 3.20GHz. 
The training of our E.T.Track architecture is based on the training framework used in LightTrack~\cite{yan2021lighttrack} which, in turn, is based on OCEAN~\cite{zhang2020ocean}.
As is common practice, we initialize the backbone with ImageNet pre-trained weights. 
The models are optimized using stochastic gradient decent~\cite{ruder2016overview} with a momentum of 0.9, and a weight decay of $1 \text{e-}4$ for 50 epochs.
During the first 10 epochs, the backbone parameters remain frozen.
We use a step learning rate scheduler during a warmup period of 5 epochs, increasing the learning rate from $2\text{e-}2$ to $1\text{e-}1$, followed by a logarithmically decreasing learning rate from $1\text{e-}1$ to $2\text{e-}4$ for the remainder. 
We utilize 3 GPUs and sample 32 image pairs per GPU for each batch. 
The sampled image pairs consist of a $256 \times 256$ search frame and a $128 \times 128$ template frame, sampled from training splits of LaSOT~\cite{fan2019lasot}, TrackingNet~\cite{muller2018trackingnet}, GOT10k~\cite{huang2019got} and COCO~\cite{lin2014microsoft}. 
Specifically, the two frames are sampled within a range of 100 frames for LaSOT~\cite{fan2019lasot} and GOT10k~\cite{huang2019got}, 30 frames for TrackingNet~\cite{muller2018trackingnet}, and 1 frame for COCO~\cite{lin2014microsoft}. 
Both patches are further shifted and scaled randomly. 

\subsection{Comparison to State-of-the-Art}
\label{sec:results}
We compare our proposed E.T.Track to state-of-the-art methods on 6 benchmarks: OTB-100~\cite{wu2013online}, NFS~\cite{kiani2017need}, UAV-123~\cite{mueller2016benchmark}, LaSOT~\cite{fan2019lasot}, TrackingNet~\cite{muller2018trackingnet}, and VOT2020~\cite{kristan2020eighth}. 
Specifically, we evaluate transformer-based trackers~\cite{transformer_tracker,transformer_meets_tracker,transformer_meets_tracker,stark}, realtime CPU trackers~\cite{danelljan2017eco,yan2021lighttrack}, as well as additional seminal trackers~\cite{danelljan2019atom,li2019siamrpn++,bhat2019learning,danelljan2020probabilistic,voigtlaender2020siam}. 
For all methods we further report the CPU runtime in \textit{FPS}.

\parsection{LaSOT~\cite{fan2019lasot}} 
The LaSOT dataset is highly challenging, and includes very long sequences with an average of 2500 frames per sequence. 
Therefore, robustness is essential to achieving a high score. 
The success plot in Fig.~\ref{fig:lasot_success} depicts the CPU realtime trackers with warmer colour continuous lines, while the non-realtime trackers are indicated by dashed lines in colder colours. 
Unlike online-learning methods such STARK that utilize a dynamically updated template, our model only uses the features of the template patch extracted in the first sequence of the frames. 
Even so, our model is very robust and reaches an AUC score of $59.1\%$, outperforming the popular DiMP tracker~\cite{bhat2019learning} by $2.2\%$. 
Compared to the lightweight mobile architecture of LT-Mobile~\cite{yan2021lighttrack}, our model improves the success score by an astonishing $7\%$ while achieving a comparable speed.

\parsection{NFS~\cite{kiani2017need}} 
We additionally evaluate our approach on the NFS dataset that contains fast moving objects.
The results are presented in Table~\ref{tab:uav_nfs_otb}. 
E.T.Track reaches an AUC score of $59\%$, outperforms all realtime trackers by at least $3.7\%$.

\parsection{OTB-100~\cite{wu2013online}} OTB-100 contains 100 sequences. As shown in Table~\ref{tab:uav_nfs_otb}, the current state-of-the-art is achieved by the recently introduced TrDiMP~\cite{transformer_meets_tracker} with  an AUC score of $71.1\%$. 
Our model achieves an AUC score of $67.8\%$, marking it as the best performing realtime tracker.

\parsection{UAV-123~\cite{mueller2016benchmark}} 
UAV-123 contains a total of 123 sequences from aerial viewpoints. 
The AUC results are shown in Table~\ref{tab:uav_nfs_otb}. 
Unlike the other datasets, E.T.Track performs comparably to LT-Mobile, with a performance of $62.3\%$.

\parsection{TrackingNet~\cite{muller2018trackingnet}} 
We further evaluate the trackers on the 511 sequences of the TrackingNet test set, and report the results  in Table~\ref{tab:trackingnet}. 
Similarly to the other datasets, E.T.Track outperforms all other realtime trackers.
Specifically, E.T.Track improves LT-Mobile precision by $1.05\%$, normalized precision by $2.42\%$, and AUC by $2.48\%$. 
Comparing E.T.Track to more complex transformer-based trackers such as TrSiam~\cite{transformer_meets_tracker}, our model is only $2.2\%$ worse in terms of precision, $2.32\%$ in terms of normalized precision, and $3.12\%$ in terms of AUC while running almost $8\times$ faster on a CPU.
This further demonstrates that, while transformers have the capability to significantly improve performance, transformer modules do not need to be prohibitively expensive for computationally constrained devices to achieve most of the performance gains.

\parsection{VOT-ST2020 \cite{kristan2020eighth}} 
Finally, we also evaluate bounding box predicting trackers on the anchor-based short term tracking dataset of VOT-ST2020. 
Unlike other tracking datasets, VOT2020 contains various anchors that are placed $\Delta_{anc}$ frames apart. 
The trackers are evaluated in terms of Accuracy, Robustness and Expected Average Overlap (EAO).
Accuracy represents a weighted combination of the average overlap between the ground truth and the predicted target predictions on subsequences defined by anchors. 
Robustness indicates the percentage of frames before the trackers fails on average. 
Finally, EAO is a measure for the overall tracking performance and combines the accuracy and the robustness. 
The results are shown in Table~\ref{tab:vot2020}. 
While our model outperforms the lighweight convolutional baseline model introduced in~\cite{yan2021lighttrack} by $1-2\%$ in terms of accuracy and robustness, the largest performance increase can be noted in terms of robustness, where the performance is increased by $5.2\%$. 
We find that learning exemplar representations from the dataset coupled with an image-level query representation significantly increases the tracker's robustness compared to its convolutional counterpart.

\subsection{Ablation Study}
\label{sec:ablation}

To further understand the contributions of the different components, we conduct a number of controlled experiments on three datasets.
Specifically, we report AUC on OTB-100~\cite{wu2013online}, NFS~\cite{kiani2017need}, and LaSOT~\cite{fan2019lasot}.

\parsection{Baseline} 
We commence our ablation study from the mobile architecture of \cite{yan2021lighttrack}, LT-Mobile, due to its performance versus efficiency trade-off.
We refer to LT-Mobile, our baseline Siamese tracker, as the Convolutional(Conv) baseline.
In LT-Mobile, the similarity of the search and template patch features is computed by a pointwise cross-correlation.
The feature map is then passed to the tracker head consisting of two branches, the classification and the bounding box regression branches, as explained in Sec.~\ref{sec:impl_details}.
The performance of the baseline model, Conv, is reported in Table~\ref{tab:ablation}.

\parsection{Exemplar Attention} 
We first evaluate the efficacy of the Exemplar Attention as a drop in replacement for the convolutional layer. 
We replace the convolutional layers with the Exemplar Attention, followed by a residual connection and a normalization layer.
In other words, setting the FFN ($f(\cdot)$) in Eq.~\ref{eq:transformer} to identify.
We report the performance of the Attention (Att) module in Table~\ref{tab:ablation}.
The performance on NFS increases by $1.3\%$, and on LaSOT by $1.5\%$, demonstrating the effectiveness of our Exemplar Attention module.
We note that this performance increase is without the use of the FFN, a key design choice in the transformer architectures~\cite{vaswani2017attention}.

\begin{table*}[t]
\centering%
\scalebox{0.8}{
\begin{tabular}{@{}cccccccc@{}}
    \toprule
   
    & Conv & Standard & Clustered & Linear & Local & Swin & E.T.Track \\
    & \cite{yan2021lighttrack} & \cite{vaswani2017attention} & \cite{vyas2020fast} & \cite{katharopoulos2020transformers} & \cite{ramachandran2019stand} & \cite{liu2021swin} & (Ours)\\
    
    \midrule
    NFS & 55.3 & 55.3 & 57.5 & 55.8 & 55.8  & 55.4 & \textcolor{blue}{59.0}\\
    OTB-100 & 66.2 & 65.3 & 67.5 & 65.4 & 64.8  & 64.2 & \textcolor{blue}{67.8} \\
    LaSOT & 52.1 & 54.2 & 56.5 & 53.5 & 53.4 & 56.9 & \textcolor{blue}{59.1}\\
    \bottomrule
 \end{tabular}}
  \caption{Comparison of the convolutional baseline (Conv) and the different attention modules in terms of AUC on NFS, OTB, and LaSOT datasets. The best score is highlighted in \textcolor{blue}{blue}. E.T.Track consistently outperforms all other transformer variants.}
 \label{tab:transformers}
\end{table*}

\parsection{FFN} Similar to the original Transformer architecture~\cite{vaswani2017attention}, we evaluate the effect of additionally using a lightweight FFN followed by a LayerNorm layer.
We find that the additional expressivity introduced by the FFN improves the performance on all three datasets, as seen in Table~\ref{tab:ablation}. 
The highest performance increase is achieved on LaSOT, where the AUC score increases by $5.5\%$. 
This yields our final E.T.Track model, depicted in Fig.~\ref{fig:tracker}. 

\parsection{Template conditioning} The queries used in the Exemplar Transformer so far are solely based on a transformed version of the initial correlation map. 
We further explore the impact of incorporating template information into our Exemplar Attention module. 
Specifically, we average pool the feature map corresponding to the template patch, and sum the representation to each layer's input. 
As seen from the Template Conditioning (T-Cond) experiments in Table~\ref{tab:ablation}, the richer queries lead to an improvement on NFS.
However, on OTB-100 and LaSOT, the model did not benefit from the additional information.
To this extend, we decide to not use the T-Cond module in our final module, keeping our final model simpler.

\begin{table}[t]
    \begin{subtable}[h]{\linewidth}
    \centering%
    \scalebox{0.8}{
        \begin{tabular}{@{}cccc|ccc@{}}
            \toprule
                Conv  & Att & FFN & T-Cond. & NFS & OTB-100 & LaSOT \\
                \midrule
                \checkmark &  &  &  & 55.3 & 66.2 & 52.1 \\
                 & \checkmark &  &  & 56.6 & 65.8 & 53.6 \\
                 & \checkmark & \checkmark &  & 58 & \textcolor{blue}{67.3} & \textcolor{blue}{59.1} \\
                 & \checkmark & \checkmark & \checkmark & \textcolor{blue}{59.0} & 66.9 & 57.9 \\
            \bottomrule
      \end{tabular}}
      \caption{Ablating the different component of the Exemplar Transformer module. 
      We evaluate the Exemplar Attention (Att) module, Feed-Forward Network (FFN), and Template Conditioning (T-Cond). The final model, depicted in Fig.~\ref{fig:tracker} includes the Att and FFN modules.}
      \label{tab:ablation}
      \vspace{0.1in}
    \end{subtable}
    \hfill
    
    \begin{subtable}[h]{\linewidth}
    \centering%
    \scalebox{0.8}{
    \begin{tabular}{@{}cccccccc@{}}
        \toprule
       
        & Conv && 1-Ex && 4-Ex && 16-Ex\\
        
        \midrule
        NFS & 55.3 && 57.6 && \textcolor{blue}{58.0} && \textcolor{blue}{58.0}\\
        OTB-100 & 66.2 && 66.5 && \textcolor{blue}{67.3} && 66.1 \\
        LaSOT & 52.1 && 57.2 && \textcolor{blue}{59.1} && 57.4\\
        \bottomrule
      \end{tabular}}
      \caption{Effect of the number of exemplars (-Ex).}
     \label{tab:n_exemplars}
     \end{subtable}
     \vspace{0.1in}
    \hfill
    
    \begin{subtable}[h]{\linewidth}
        \centering%
        	\scalebox{0.8}{
        \begin{tabular}{@{}c|cc|cc|cc|cc@{}}
            \toprule
           
            & \multicolumn{2}{c|}{ShuffleNet} & \multicolumn{2}{c|}{MobileNetV3} & \multicolumn{2}{c|}{ResNet-18} & \multicolumn{2}{c}{LT-Mobile} \\
            
            & \multicolumn{2}{c|}{\cite{zhang2018shufflenet}} & \multicolumn{2}{c|}{\cite{howard2019mobilenet}} & \multicolumn{2}{c|}{\cite{he2016deep}} & \multicolumn{2}{c}{\cite{yan2021lighttrack}} \\
        
            \midrule
            Conv & \checkmark &  & \checkmark &  & \checkmark &  & \checkmark &   \\
            E.T. (Ours) &   & \checkmark &  & \checkmark &  & \checkmark &  & \checkmark \\
            \midrule
            
            NFS & 54.9 & \textcolor{blue}{56.2} & \textcolor{blue}{56.8} & \textcolor{blue}{56.8} & 55.8 & \textcolor{blue}{57.3} & 55.3 & \textcolor{blue}{59.0}\\
            OTB-100 & 61.3 & \textcolor{blue}{61.8} & 64.5 & \textcolor{blue}{65.3} & 65.3 & \textcolor{blue}{65.7} & 66.2 & \textcolor{blue}{67.8}\\
            LaSOT & 48.6 & \textcolor{blue}{49.8} & 52.1 & \textcolor{blue}{52.7} & 55.9 & \textcolor{blue}{56.5} & 52.1 & \textcolor{blue}{59.1}\\
            \bottomrule
          \end{tabular}}
          \caption{Comparison of Exemplar Transformer (E.T.) and Convolutional (Conv) modules on different backbone models. The E.T. consistently outperforms the Conv models, independently of the backbone used.}
         \label{tab:backbones}
     \end{subtable}
     \caption{Ablation experiments reported in terms of AUC on NFS, OTB, and LaSOT datasets. Conv refers to LT-Mobile~\cite{yan2021lighttrack} that acts as our convolutional baseline. The best score is highlighted in \textcolor{blue}{blue}.}
     \label{tab:ablations}
     \vspace{-0.2in}
\end{table}

\parsection{Number of Exemplars}
Table~\ref{tab:n_exemplars} reports the performance given the number of Exemplars. 
While more Exemplars increases the overall capacity of the model, and as such, one would expect further performance gains, our experiments yield different results.
Specifically, 4 Exemplars yield consistently better results across all the datasets. 
We hypothesize that, while training a model with a larger number of experts can increase performance, modifications in the optimization process are required to ensure the selection of the appropriate exemplar.
The efficient implementation of the Exemplar Attention module, Eq.~\ref{eq:et-final_light}, ensures a comparable runtime even with a larger number of exemplars.

Interestingly, while single Exemplar Attention is mathematical equivalent to a regular convolution with a residual operation, the additional FFN layer following the Exemplar Attention increases considerably the performance. 
Specifically, we observe a performance increase of 2.3\% on NFS, 0.3\% on OTB-100, and 5.1\% on LaSOT.

\parsection{Backbone Alternatives}
All experiments reported so far utilize the LT-Mobile encoder. 
To demonstrate the flexibility of Exemplar Transformers, as well as their independence to the encoder architecture, we evaluate the use of different encoder architectures.
Specifically, we compare the performance of the two tracker head module variants (Convolution, Exemplar Transformer) in combination with ShuffleNet~\cite{zhang2018shufflenet}, MobileNetV3~\cite{howard2019mobilenet}, ResNet-18~\cite{he2016deep}, and LT-Mobile~\cite{yan2021lighttrack}. 
The results presented in Table~\ref{tab:backbones} demonstrate consistent performance gains independent of the encoder architecture, highlighting the superiority of our Exemplar Transformer to its convolutional counterpart.

\parsection{Comparison of Alternative Transformer Layers}
To validate the design choices and hypothesis that lead to the Exemplar Transformer module, we additionally compare to other Transformer Layer variants.
All Transformer layers evaluated can also act as drop-in replacements to standard convolutions.
Specifically, we evaluate the Standard~\cite{vaswani2017attention}, Clustered~\cite{vyas2020fast}, Linear~\cite{katharopoulos2020transformers}, Local~\cite{ramachandran2019stand}, and Swin~\cite{liu2021swin} Transformers.
The selection ensures at least one method from every transformer category defined in Sec.~\ref{sec:related_work}, while using their official public implementations ensures a fair comparison.
The results in Table~\ref{tab:transformers} demonstrate that our Exemplar Transformer (E.T.) consistently outperforms all other attention variants across all the datasets.
These findings further validate our hypothesis that one global query and a small set of exemplar representations is sufficiently descriptive when tracking a single object.

\label{sec:experiments}

\section{Conclusion}
We propose a novel Transformer layer for single object visual tracking, based on Exemplar Attention. 
Exemplar Attention utilizes a single query token of the input sequence, and jointly learns a small set of exemplar representations. 
The proposed transformer layer can be used throughout the architecture, \eg as a substitute for a convolutional layer. 
Having a comparable computationally complexity to standard convolutional layers while being more expressive, the proposed Exemplar Transformer layers can significantly improve the accuracy and robustness of tracking models with minimal impact on the model's overall runtime. 
E.T.Track, our Siamese tracker with Exemplar Transformer, significantly improve the performance compared to the convolutional baseline and other transformer variants. 
E.T.Track is capable of running in realtime on computationally limited devices such as standard CPUs.

\label{sec:conclusion}

\newpage

\renewcommand\thefigure{S.\arabic{figure}}    
\renewcommand\thetable{S.\arabic{table}}   

\begin{center}
	\textbf{\Large Supplementary Material}
\end{center}

In this supplementary material, we first provide an ablation study of the number of query vectors $S^2$ in Sec.~\ref{appendix:s_ablation}. 
We present the pseudocode of our Exemplar Transformer Layer in Sec.~\ref{appendix:algo}. 
We provide additional results on the VOT2020~\cite{kristan2020eighth} real-time challenge in Sec.~\ref{appendix:vot}. 
In Sec.~\ref{appendix:video} we analyze the qualitative results of our tracker on a selection of sequences, while in Sec.~\ref{appendix:attributes} we further examine the results of our tracker on the LaSOT dataset~\cite{fan2019lasot} with respect to the specific attributes. 
Finally, we present the success plots of the NFS~\cite{kiani2017need}, OTB-100~\cite{wu2013online}, and UAV-123~\cite{mueller2016benchmark} datasets in Sec.~\ref{appendix:success}. 
The code and the instructions to reproduce our results are included in the supplementary material folder, and will be made available upon publication.
\appendix

\section{Ablation of Number of Query Vectors}
 
As explained in 
Sec.~\ref{sec:eff_tracker}
of the main paper, we set $S=1$ in our experiments based on the assumption that one global token encapsulates sufficient information for the task of single object tracking. 
To further evaluate this hypothesis, we ablate the parameter $S$. 
Specifically, the input feature map is divided into $S \times S$ patches, for which we compute individual query vectors.
Table~\ref{supp:tab:s_ablation} presents the results of our experiments. 
The results confirm our assumption that utilizing a single token as global representation yields the best results.

\begin{table}[h]
    \centering%
    	\scalebox{0.9}{
  \begin{tabular}{@{}cccccc@{}}
        \toprule
    & S=1 && S=2 && S=4\\
        \midrule
    NFS & \textcolor{blue}{59.0} && 46.6 && 46.7 \\
    OTB-100 & \textcolor{blue}{67.8} && 55.5 && 57.5 \\
    LaSOT & \textcolor{blue}{59.1} && 43.7 && 42.6 \\

        \bottomrule
      \end{tabular}}
\caption{Ablation experiment of the different values for $S$ reported in terms of AUC on NFS, OTB, and LaSOT datasets. 
Utilizing a global query token ($S=1$) yields consistently better results. 
The best score is highlighted in \textcolor{blue}{blue}.
}  
     \label{supp:tab:s_ablation}
 \end{table}
\label{appendix:s_ablation}

\section{Algorithm}

We present below the pseudocode for the Exemplar Attention layer 
(Eq.~\ref{eq:et-final_light})
, depicted on the right side of 
Fig.~\ref{fig:att_module}
in Algorithm~\ref{sup:algo:ET}. 

\begin{algorithm}
    \caption{Pseudocode of the Exemplar Attention layer, 
    Eq.~\ref{eq:et-final_light}.} 
    \DontPrintSemicolon
    \SetKwFunction{ExemplarAttention}{ExemplarAttention}
    \SetKwProg{Fn}{function}{:}{end}
    \Fn{\ExemplarAttention{${X}$}}{
        $Q \gets \Psi_S(X)W_Q$ \hfill Eq.~
        \ref{eq:avg_pool}\\
        $\hat{K} \gets \hat{W}_K$ \\
        $\hat{V} \gets W_V$ \\
        $\text{sim} \gets \text{softmax}({Q \cdot \hat{K}^T)}$  \\
        $\text{sim} \gets \text{sim} / \sqrt{d_k}$ \\
        $W_A = \text{sim} \cdot \hat{V}$ \\
        $A(X) \gets W_A \circledast X$ \\

        \KwRet $A(X)$
    }
\label{sup:algo:ET}
\end{algorithm}

\label{appendix:algo}

\section{VOT-RT2020}

We evaluate bounding box predicting trackers on the anchor-based short term tracking dataset of VOT-RT2020~\cite{kristan2020eighth}, similar to 
Sec.~\ref{sec:results}. 
The results are presented in Table~\ref{sup:tab:vot2020}. 
While the performance of our model is comparable to LT-Mobile~\cite{yan2021lighttrack} in terms of accuracy, our model is nearly $6\%$ better in terms of robustness. 
We find that learning exemplar representations from the dataset coupled with an image-level query representation significantly increases the tracker's robustness compared to its convolutional counterpart.

\begin{table*}[t]
  \centering%
\scalebox{0.9}{
  \begin{tabular}{@{}cggggccc@{}}
    \toprule
    & \multicolumn{4}{g}{non-realtime} & \multicolumn{3}{c}{realtime} \\
    \midrule
    & SiamFC & ATOM & DiMP & SuperDiMP &  KCF & LT-Mobile & \textbf{E.T.Track} \\
    & \cite{bertinetto2016fully} & \cite{danelljan2019atom} & \cite{bhat2019learning} & \cite{superdimp} & \cite{henriques2014high} & \cite{yan2021lighttrack} & \textbf{(Ours)} \\
    \midrule
    EAO & 0.172 & 0.237 & 0.241 & \textcolor{blue}{0.289}  & 0.154 &  0.217 & \textbf{\textcolor{red}{0.227}}\\
    Accuracy & 0.422 & 0.440 & 0.434 & \textcolor{blue}{0.472} & 0.406 &  0.418 & \textbf{\textcolor{red}{0.418}}\\
    Robustness & 0.479 & 0.687 & 0.700 & \textcolor{blue}{0.767} & 0.434 & 0.607 & \textbf{\textcolor{red}{0.663}}\\
    \midrule
    CPU Speed  & 6 & 20 & 15 & 15 & 95 & 47 & 47 \\
    \bottomrule
  \end{tabular}}
  \caption{Comparison of bounding box predicting trackers on the VOT-RT2020 dataset. 
  We report the Expected Average Overlap (EAO), the Accuracy and Robustness.
  The best score is highlighted in \textcolor{blue}{blue} while the best realtime score is highlighted in \textcolor{red}{red}. We additionally report CPU runtime speeds in \textit{FPS}.} 
 \label{sup:tab:vot2020}
\end{table*}

\label{appendix:vot}

\section{Video Visualizations}
We additionally provide sequence comparisons between E.T.Track and LT-Mobile~\cite{yan2021lighttrack}.
Table~\ref{tab:videos} lists the sequences compared, and reports their performance.
In addition, the associated videos can be found in the supplementary folder. 

\parsection{person8-2}
The person8-2 sequence of the UAV-123 dataset~\cite{mueller2016benchmark} of a man running on grass nicely demonstrates that our tracker does not lose track of the target even when he partially moves out of the frame.
Specifically, E.T.Track is able to completely recover when the target moves back into the frame. LT-Mobile~\cite{yan2021lighttrack} yields comparable results.

\parsection{Human7}
Human7 from the OTB dataset~\cite{wu2013online} films a woman walking. 
Even though the video appears to be jittery, the appearance and shape of the target object changes only marginally. 
Our model achieves an average overlap of 88\% which is 7\% higher than LT-Mobile~\cite{yan2021lighttrack}.

\parsection{boat-9}
The boat-9 from the UAV-123 dataset~\cite{mueller2016benchmark} depicts a target which not only changes appearance, but also significantly decreases in size due to an increasing distance to the camera.
We find that E.T.Track can still handle such scenarios, and unlike LT-Mobile, it maintains track of the boat even after a $180$-degree turn.
E.T.Track is therefore more robust than LT-Mobile, attributed to the increased capacity introduced by the Exemplar Transformer layers.

\parsection{basketball-3}
In the basketball-3 sequence of NFS~\cite{kiani2017need}, the increased robustness introduced by the Exemplar Transformer layer enables the separation between the player's head and the basketball, unlike LT-Mobile.

\parsection{drone-2}
The drone-2 sequence of LaSOT~\cite{fan2019lasot} shows a target that shortly moves completely out of the frame, and later re-enters the scene with a different appearance to the initial frame. 
Furthermore, the target object's location deviates from the tracker's search range when re-entering the scene. 
These two aspects pose a challenge both for our model, as well as LT-Mobile~\cite{yan2021lighttrack}, and are inherent limitations of the tracking inference pipeline used in both approaches~\cite{zhang2020ocean}. 
Specifically, the tracking pipeline contains a post-processing step in which the predicted bounding boxes are refined. 
Changes in size, as well as changes of the bounding box aspect ratios, are therefore penalized. 
In addition, both models search only within a small image patch around the previously predicted target location. 
This challenge can potentially be addressed by integrating our Exemplar Transformer layer into trackers that directly predicts bounding boxes without any post-processing.
We did not investigate this further, but consider this an interesting direction for future research. 

\begin{table}
  \centering
\scalebox{0.9}{
  \begin{tabular}{@{}lccc@{}}
    \toprule
    & Dataset & LT-Mobile & \textbf{E.T.Track} \\
    & & \cite{yan2021lighttrack} & \textbf{(Ours)}\\
    \midrule
    person8-2 & UAV-123~\cite{mueller2016benchmark} & 0.889 & \textbf{\textcolor{blue}{0.915}} \\
    Human7 & OTB~\cite{wu2013online}  & 0.813 & \textbf{\textcolor{blue}{0.883}}\\
    boat-9 & UAV-123~\cite{mueller2016benchmark}  & 0.483 & \textbf{\textcolor{blue}{0.803}} \\
    basketball-3 & NFS~\cite{kiani2017need}  & 0.259 & \textbf{\textcolor{blue}{0.707}} \\
    drone-2 & LaSOT~\cite{fan2019lasot} & 0.192 & \textbf{\textcolor{blue}{0.887}} \\
    \bottomrule
  \end{tabular}}
  \caption{Direct per-sequence comparison of E.T.Track and LT-Mobile~\cite{yan2021lighttrack} on various sequences in terms of Average Overlap (AO). The best performance is highlighted in \textcolor{blue}{blue}.}
 \label{tab:videos}
\end{table}
\label{appendix:video}

\section{Attributes}

Table~\ref{sup:tab:lasot_attributes} presents the results of various trackers on different sequence attributes of the LaSOT dataset~\cite{fan2019lasot}. 
We consistently outperform the other realtime trackers by a significant margin in every attribute. 
The attribute with the largest performance gains compared to LT-Mobile~\cite{yan2021lighttrack} are \emph{Full Occlusion} with $10.4\%$, \emph{Motion Blur} with $9.7\%$, \emph{Background Clutter} with $8.5\%$, and \emph{Fast Motion} with $7.5\%$.
These attributes are either known limitation of the tracking pipeline utilized~\cite{zhang2020ocean}, as discussed in Sec.~\ref{appendix:video}, or can benefit from increased network capacity.
We find that the incorporation of our Exemplar Transformer layers increases robustness and improves attributes that are even known limitations of the overall framework.

When comparing our model to the non-realtime state-of-the-art STARK~\cite{stark}, our model observes an average performance drop of $-7.3\%$.
The most challenging attributes are \emph{Viewpoint Change}, \emph{Full Occlusion}, \emph{Fast Motion}, \emph{Out-of-View}, and \emph{Low Resolution}.
This analysis paves the path for future research in the design of novel modules for efficient tracking, specifically tackling the identified challenging attributes.

\begin{table*}[t]
	\centering
	\newcommand{\best}[1]{\textbf{\textcolor{red}{#1}}}
	\newcommand{\scnd}[1]{\textbf{\textcolor{blue}{#1}}}
	\newcommand{\dist}{\hspace{4pt}}%
	\resizebox{\textwidth}{!}{%
        \begin{tabular}{l@{\dist}c@{\dist}c@{\dist}c@{\dist}c@{\dist}c@{\dist}c@{\dist}c@{\dist}c@{\dist}c@{\dist}c@{\dist}c@{\dist}c@{\dist}c@{\dist}c@{\dist}|c@{\dist}}
        	\toprule
        	                       & Illumination & Partial     &                & Motion         & Camera      &             & Background & Viewpoint   & Scale        & Full        & Fast        &             & Low         & Aspect        &       \\
        	                       & Variation    & Occlusion   & Deformation    & Blur           & Motion      & Rotation    & Clutter    & Change      & Variation    & Occlusion   & Motion      & Out-of-View & Resolution  & Ration Change & Total \\
 
        	\midrule
        	\rowcolor{Gray}
        	STARK-ST50      & 66.8                      & \textcolor{blue}{64.3}                 & 66.9           & 62.9           & \textcolor{blue}{69.0}             & \textcolor{blue}{66.1}        & 57.3                  & \textcolor{blue}{67.8}                & \textcolor{blue}{66.1}            & \textcolor{blue}{58.7}              & \textcolor{blue}{53.8}           & \textcolor{blue}{62.1}           & \textcolor{blue}{59.4}              & \textcolor{blue}{64.9}   & \textcolor{blue}{66.4} \\                
\rowcolor{Gray}
TransT          & 65.2                      & 62.0                 & \textcolor{blue}{67.0}           & \textcolor{blue}{63.0}           & 67.2             & 64.3        & 57.9                  & 61.7                & 64.6               & 55.3              & 51.0           & 58.2           & 56.4              & 63.2   & 64.9 \\                
\rowcolor{Gray}
TrDiMP          & \textcolor{blue}{67.5}                      & 61.1                 & 64.4           & 62.4           & 68.1             & 62.4        & \textcolor{blue}{58.9}                  & 62.8                & 63.4               & 56.4              & 53.0           & 60.7           & 58.1              & 62.3   & 63.9 \\                
\rowcolor{Gray}
TrSiam          & 63.8                      & 60.1                 & 63.8           & 61.1           & 65.5             & 62.0        & 55.1                  & 60.8                & 62.5               & 54.5              & 50.6           & 58.9           & 56.0              & 61.2   & 62.6 \\                
\rowcolor{Gray}
PrDiMP50        & 63.3                      & 57.1                 & 61.3           & 58.0           & 64.0             & 59.1        & 55.4                  & 61.7                & 60.1               & 51.6              & 49.2           & 57.0           & 54.8              & 59.0   & 60.5 \\   
\rowcolor{Gray}
DiMP            & 59.5                      & 52.1                 & 56.6           & 54.6           & 59.3             & 54.5        & 49.7                  & 56.7                & 55.8               & 47.5              & 45.6           & 49.5           & 49.1              & 54.5   & 56.0 \\                
\rowcolor{Gray}
SiamRPN++       & 53.0                      & 46.6                 & 52.8           & 44.2           & 51.3             & 48.5        & 44.9                  & 44.4                & 49.4               & 36.6              & 31.6           & 41.6           & 38.5              & 47.2   & 49.5 \\  
LT-Mobile       & 55.0                      & 48.9                 & 57.3           & 45.8           & 52.9             & 51.2        & 43.3                  & 49.9                & 51.9               & 38.3              & 33.6           & 43.7           & 40.8              & 49.9   & 52.1 \\                
SiamFC          & 34.6                      & 30.6                 & 35.1           & 30.8           & 33.3             & 31.0        & 30.8                  & 28.6                & 33.2               & 24.5              & 19.5           & 25.6           & 25.2              & 30.8   & 33.6 \\     
\textbf{E.T.Track (Ours)}       & \best{61.3}                      & \best{55.7}             & \best{61.0}           & \best{55.5}           & \best{60.2}             & \best{58.1}        & \best{51.8}                  & \best{55.9}                & \best{58.8}               & \best{48.7}              & \best{41.1}           & \best{51.1}           & \best{48.8}              & \best{56.9}   & \best{59.1} \\

            \bottomrule
        \end{tabular}
	}
	\caption{LaSOT attribute-based analysis. Each column corresponds to the results computed on all sequences in the dataset with the corresponding attribute. The trackers that do not run in real-time are highlighted in grey. The overall best score is highlighted in \textcolor{blue}{blue} while the best realtime score is highlighted in \textcolor{red}{red}.}
	\label{sup:tab:lasot_attributes}%
\end{table*}

\begin{figure*}[t]
    \centering
        \begin{subfigure}{0.3\linewidth}
            \includegraphics[width=\textwidth]{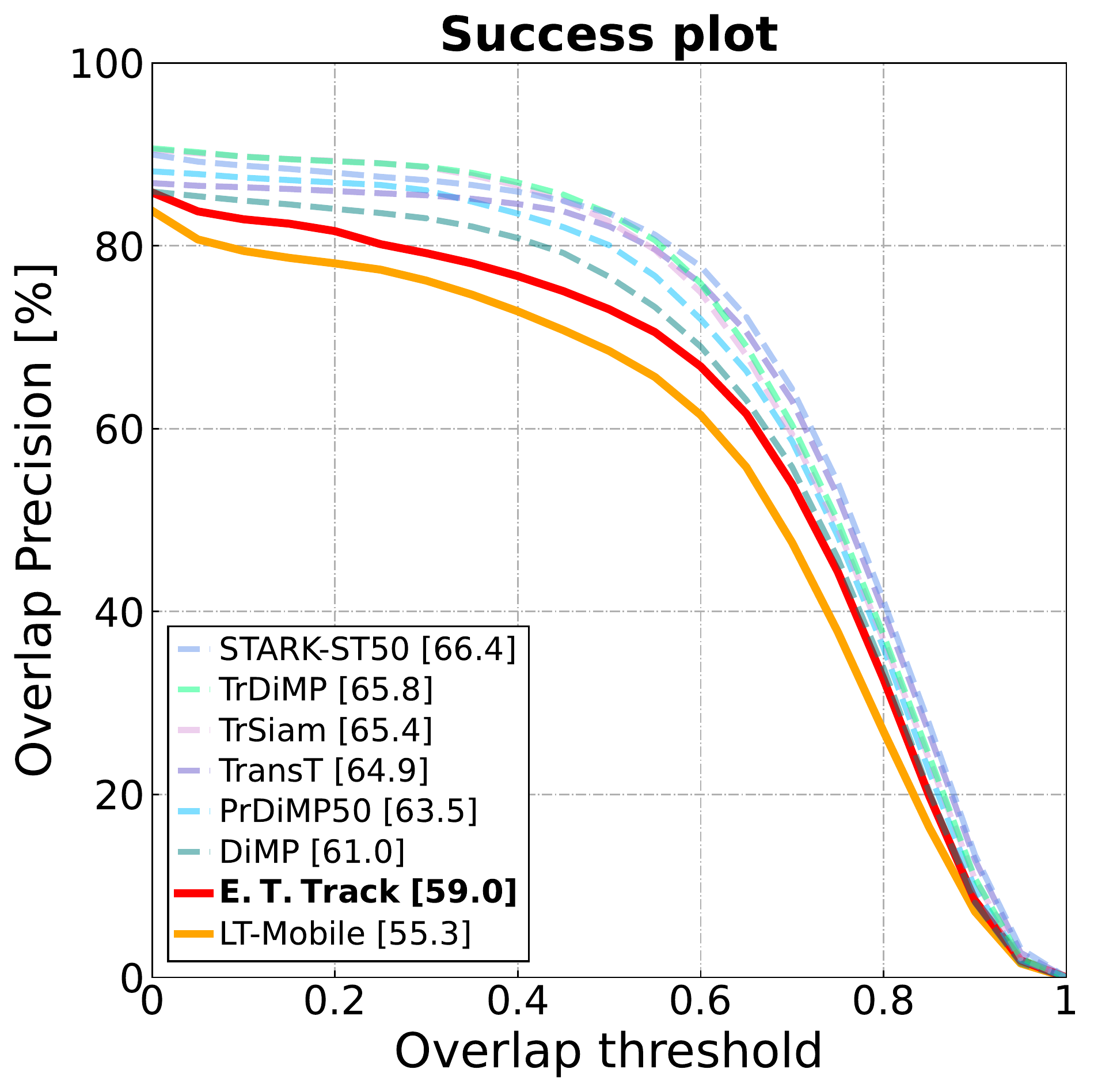}
            \caption{Success plot on the NFS dataset. Our tracker outperforms LT-Mobile~\cite{yan2021lighttrack} by a significant margin.}
            \label{sup:fig:nfs}
        \end{subfigure}
    \hfill
        \begin{subfigure}{0.3\linewidth}
            \includegraphics[width=\textwidth]{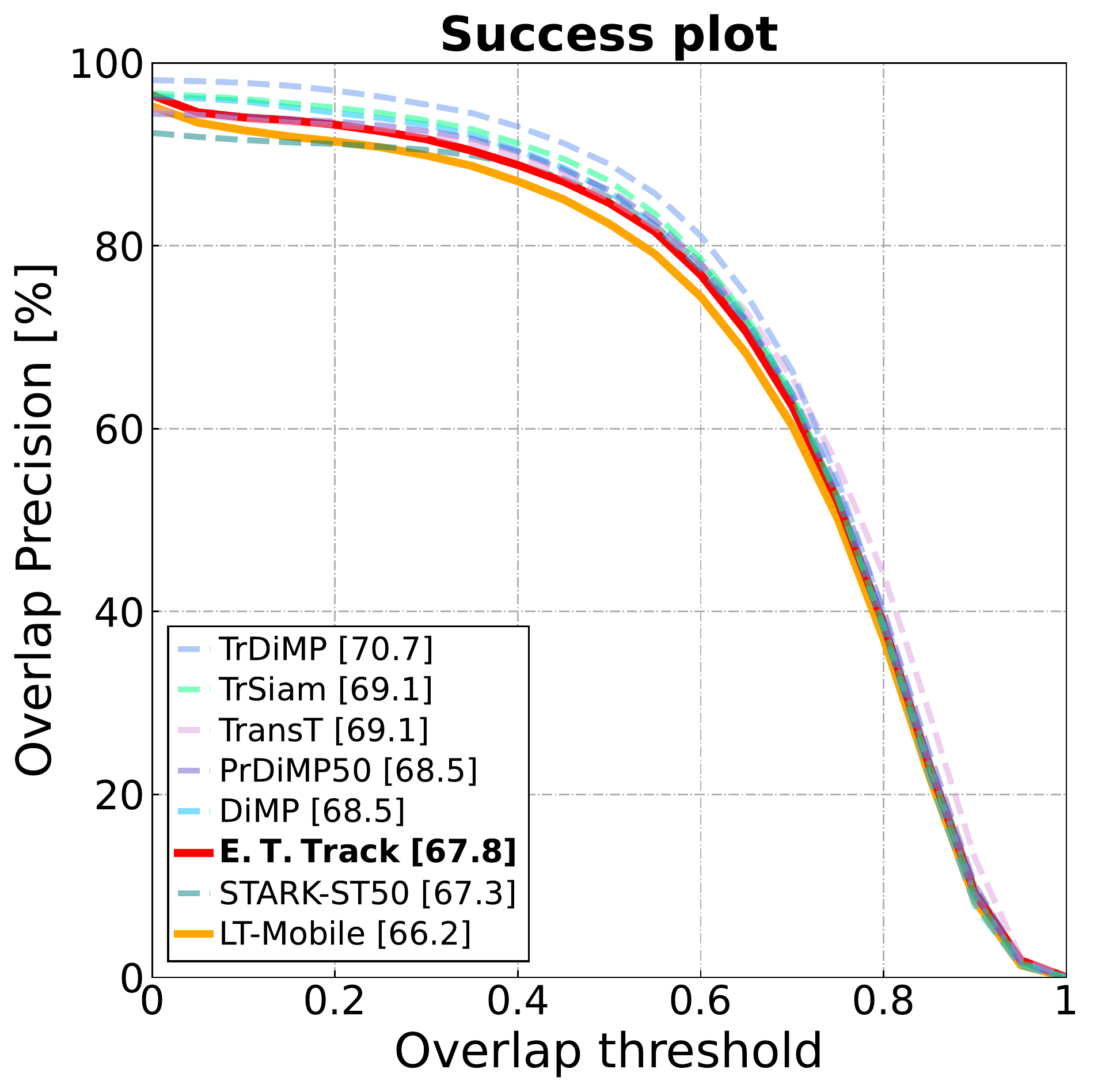}
            \caption{Success plot on the OTB-100 dataset. Our tracker outperforms LT-Mobile~\cite{yan2021lighttrack} by a small margin.}
            \label{sup:fig:otb}
        \end{subfigure}
    \hfill
        \begin{subfigure}{0.3\linewidth}
            \includegraphics[width=\textwidth]{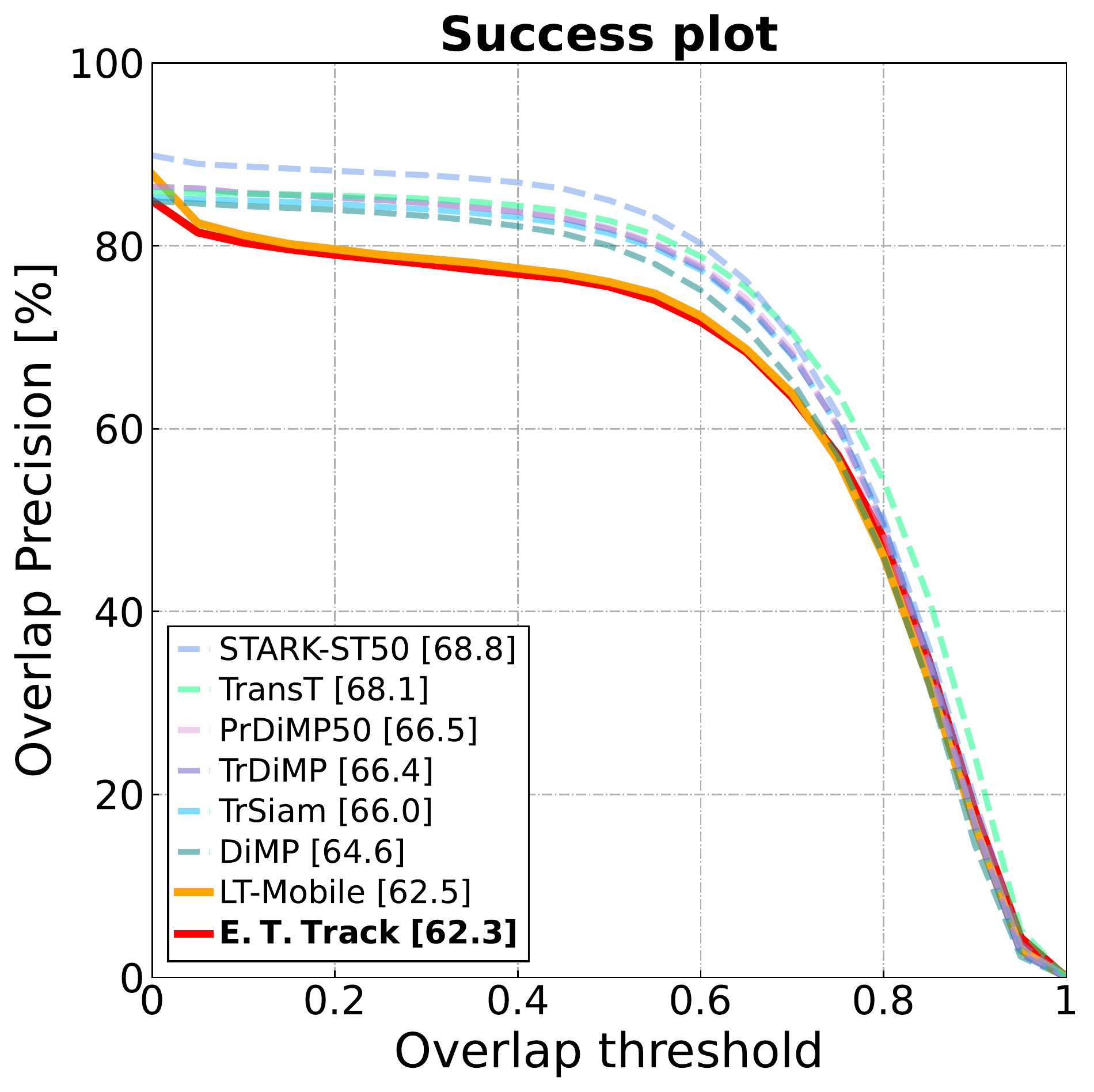}
            \caption{Success plot on the UAV-123 dataset. The performance of our tracker is comparable to the performance of LT-Mobile~\cite{yan2021lighttrack}.}
            \label{sup:fig:uav}
        \end{subfigure}
    \caption{Success plots. The CPU realtime trackers are indicated by continuous lines in warmer colours, while the non-realtime trackers are indicated by dashed lines in colder colours.}
    \label{sup:fig:success}
    \vspace{3.3in}
\end{figure*}

\label{appendix:attributes}

\section{Additional Success Plots}

We depict the success plot of the NFS dataset~\cite{kiani2017need} in Fig.~\ref{sup:fig:nfs}, the success plot of the OTB-100 dataset~\cite{wu2013online} in Fig.~\ref{sup:fig:otb} and the success plot of the UAV-123 dataset~\cite{mueller2016benchmark} in Fig.~\ref{sup:fig:uav}. 
For efficient trackers, we limited our comparison to the mobile architecture presented in LightTrack~\cite{yan2021lighttrack}, as SiamRPN++~\cite{li2019siamrpn++} and SiamFC~\cite{bertinetto2016fully} were consistently outperformed by a large margin.
We additionally report the non-realtime transformer-based trackers STARK-ST50~\cite{stark}, TrDimp~\cite{transformer_meets_tracker}, TrSiam~\cite{transformer_meets_tracker} and TransT~\cite{transformer_tracker}, as well as the seminal trackers DiMP~\cite{bhat2019learning} and PrDiMP~\cite{danelljan2020probabilistic}. 

The results presented in Fig.~\ref{sup:fig:success} correspond to our evaluation results, and therefore deviate slightly from those reported in 
Sec.~\ref{sec:results}
as those were directly acquired from their respective papers. 
As it can be seen, our model outperforms LT-Mobile~\cite{yan2021lighttrack} on all but one benchmark dataset. 
More importantly, we want to highlight the shrinking gap between the complex transformer-based trackers and our realtime CPU tracker. 
Closing this gap even further while maintaining the realtime speed will be a crucial part for future work in order to deploy high-performing trackers on computationally limited edge devices.

\label{appendix:success}

\newpage

{\small
\bibliographystyle{ieee_fullname}
\bibliography{bib}
}

\end{document}